%% file: main.tex
\documentclass[letterpaper, 10pt, conference]{ieeeconf}      

\IEEEoverridecommandlockouts                              
\usepackage{booktabs} 
\usepackage{graphicx}
\usepackage{subcaption}
\usepackage{mathtools}
\usepackage{amsmath}

\title{\LARGE \bf
3D Soil Compaction Mapping through \\
Kriging-based Exploration with a Mobile Robot
}

\author{Jaime Pulido Fentanes$^{1}$, Iain Gould$^{2}$, Tom Duckett$^{1}$, Simon Pearson$^{2}$ and Grzegorz Cielniak$^{1}$\\
{\tt\small \{jpulidofentanes,igould,tduckett,spearson,gcielniak\}@lincoln.ac.uk}
\thanks{*This work was supported by the STFC Newton Fund programme, project ST/N006836/1.}
\thanks{$^{1}$Lincoln Centre for Autonomous Systems, School of Computer Science, University of Lincoln, Brayford Campus, LN6 7TS Lincoln, UK}%
\thanks{$^{2}$Lincoln Institute for Agri-food Technology, University of Lincoln, Riseholme Park, LN2 2LG Lincoln, UK}%
}

\begin{document}

\maketitle
\thispagestyle{empty}
\pagestyle{empty}

\begin{abstract}

This paper presents an automated method for creating spatial maps of soil condition with an outdoor mobile robot. Effective soil mapping on farms can enhance yields, reduce inputs and help protect the environment. Traditionally, data are collected manually at an arbitrary set of locations, then soil maps are constructed offline using Kriging, a form of Gaussian process regression. This process is laborious and costly, limiting the quality and resolution of the resulting information. 
Instead, we propose to use an outdoor mobile robot for automatic collection of soil condition data, building soil maps online and also adapting the robot's exploration strategy on-the-fly based on the current quality of the map. We show how using Kriging variance as a reward function for robotic exploration allows for both more efficient data collection and better soil models. This work presents the theoretical foundations for our proposal and an experimental comparison of exploration strategies using soil compaction data from a field generated with a mobile robot.
\end{abstract}

\input{src/introduction}

\input{src/related}

\input{src/method}

\input{src/experimental}

\input{src/results}

\section{CONCLUSIONS}\label{sec:conclusions}

This work presents a method for soil compaction mapping using Kriging variance as the driving force for robotic exploration. Our results show that Kriging variance is closely correlated with the model error and that using a sampling strategy that reduces the model variance can lead to higher quality models requiring a lower number of samples. These results have been validated using a surrogate model that has been constructed from high density 3D soil compaction data captured by an automated penetrometer fitted on a mobile robot specially designed for agricultural settings.

This work presents a comparison of this surrogate model with a model created from manually collected data using state-of-the-art tools for soil compaction mapping used by soil scientists nowadays. Finally, this work provides a comparison between multiple well-known strategies for robotic exploration and adaptive sampling techniques, and discusses their performance and main characteristics.

We believe that this approach can be used in multiple soil surveyance applications to measure and map various soil variables including moisture, chemistry and biology either separately or simultaneously together. This will lead to high definition soil data which will allow for better understanding of soil, its properties and influence on agricultural processes. Finally, we are currently working on developing techniques for exploring agricultural environments where there are navigational limitations such as rows in arable fields.

\addtolength{\textheight}{-14.5cm}   







\bibliographystyle{IEEEtran}
\bibliography{main}

\end{document}

%% file: src/introduction.tex
\section{INTRODUCTION} \label{sec:intro}

A rising global population places challenges on the production capacity of our agricultural systems~\cite{foley2011solutions}. The condition of our soils is fundamental to this production, however soil quality is in decline globally~\cite{fao2015status}. To reverse this trend, effective monitoring of the condition of soils, as part of routine soil management~\cite{batey2009soil}, provides us with a better understanding of the problems we face and the steps needed to mitigate. 

One key driver of soil degradation proving to be a challenge for agricultural productivity in Europe is soil compaction, caused by compressive forces from farm machinery or livestock traffic ~\cite{DEFRA2009}. Compaction can lead to reductions in crop yields, and a need for greater fertiliser and fuel inputs, resulting in lower farm productivity and revenue~\cite{chamen2015mitigating,lipiec2003quantification} and its intensity can vary across a single field, contributing to spatial variability in yields~\cite{batey2009soil}. Mitigation options, such as subsoiling, are available to address compaction, however over the whole field scale this can incur large fuel costs. A targeted approach would prove much more cost effective~\cite{chamen2015mitigating}, but requires knowledge of spatial variability to identify compaction ‘hotspots’ within a field. As such, effective mapping of soil compaction is essential to address such issues whilst minimising remediation input costs.

\begin{figure}[t]
	\centering
	\begin{subfigure}[b]{0.48\columnwidth}
      \includegraphics[width=\columnwidth]{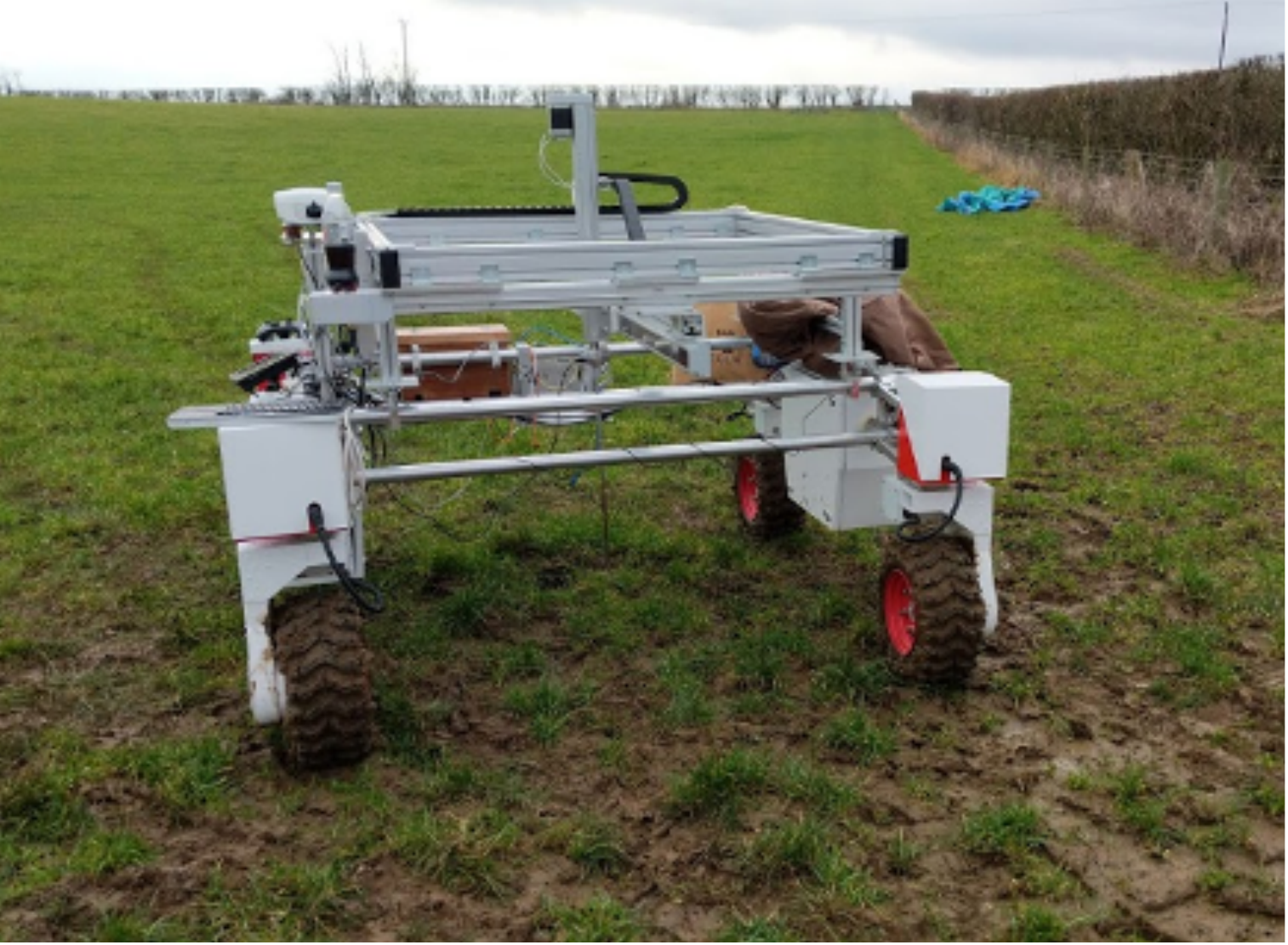}
      \caption{\label{fig:thorvald_infield}}
    \end{subfigure}
	~
    \begin{subfigure}[b]{0.47\columnwidth}
      \includegraphics[width=\columnwidth]{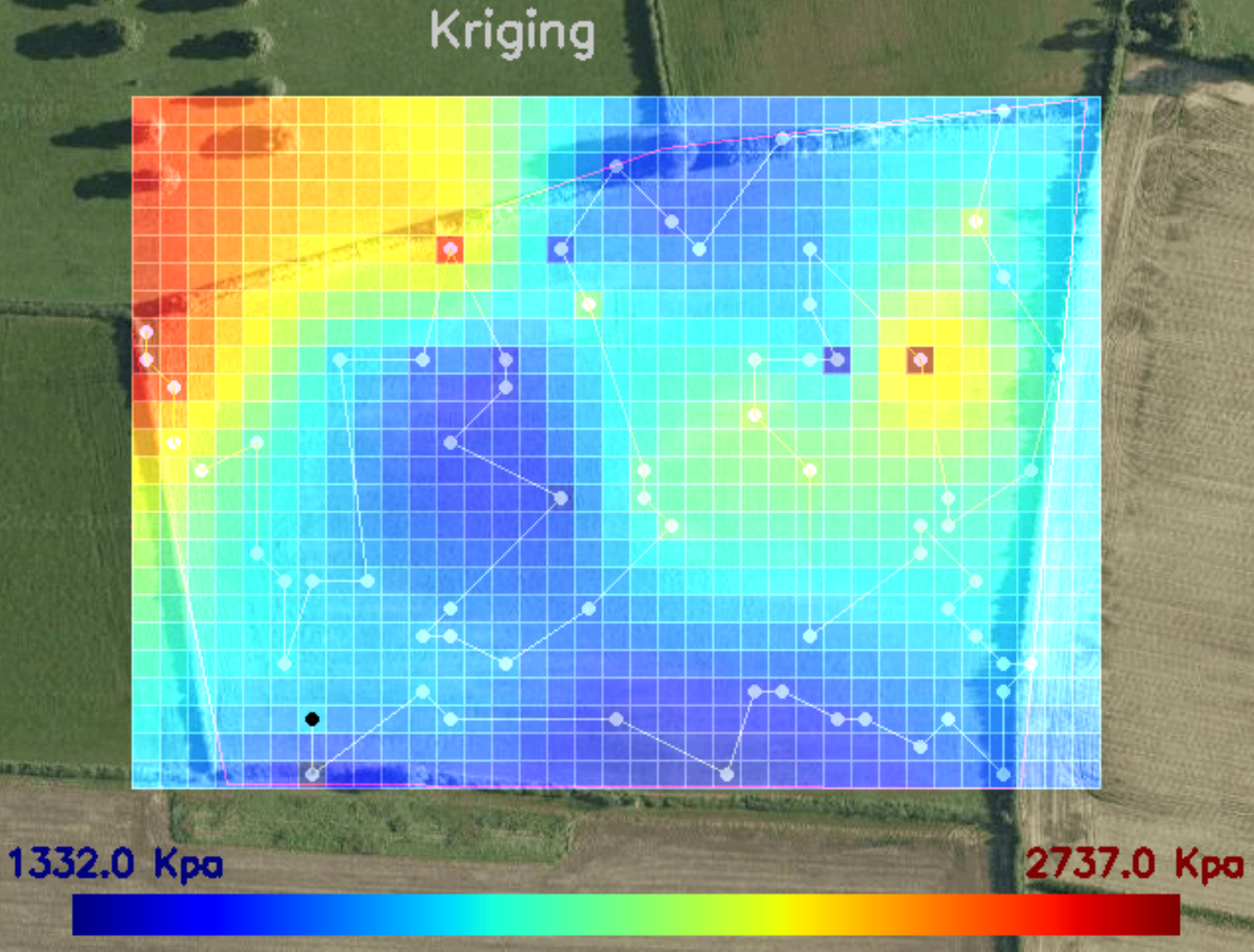}
      \caption{\label{fig:model}}
    \end{subfigure}
    \caption{(a) The outdoor mobile robot Thorvald equipped with an automated penetrometer during field experiments in January 2018. (b) A 2D slice of the 3D soil compaction map at specific depth resulting from the proposed exploration process.}
\end{figure}

Until now robotics research in agriculture has mainly focused on improving crop operations by automating tasks that demand high amounts of energy and labour, or making farm machinery more efficient and precise. 
However, mobile robots are able to create accurate models of their environment and are exceptionally good at detecting anomalies and identifying patterns. Soil is a medium which displays great variation across small spatial scales, and as such, it seems natural to use robots to measure field variability and create detailed maps of soil condition. 
To realise this vision,  \cite{bonirob2014,grmistad17thorvald} presented multi-purpose agricultural robots that can carry out different tasks in the field depending on the set of tools they are fitted with. These tasks are not limited to farming operations, but also soil surveying or maintenance. 

In this work we present a 3D soil compaction mapping application using the agricultural robot presented in \cite{grmistad17thorvald}. 
We fitted this robot with a penetrometer (see Fig.~\ref{fig:thorvald_infield}), a favoured tool for identifying spatial variability in soil compaction~\cite{hatley2005soil}, which can be used relatively quickly and with easy repeatability, making it a suitable mechanism for robotic applications. 
Scholz et al.\ previously fitted the robot presented in~\cite{bonirob2014} with a penetrometer in order to collect data from a field. Their work demonstrates how the use of robots and robotised tools is particularly beneficial for this application, and provides an excellent comparison of the accuracy of a robotised tool against manual tools. 
While their approach used a pre-determined, fixed exploration strategy, this paper investigates alternative exploration strategies which can be adapted on-the-fly depending on the current state of the soil compaction map.

The objective of this paper is to develop a methodology to generate accurate 3D soil compaction maps
with a minimal number of data samples.
To generate a high density map we use Kriging or Gaussian Process Regression~\cite{matheron1963principles}, a method widely used in geo-statistics for estimating or approximating unknown functions from data by means of interpolation.
One of the advantages of Kriging as an interpolation method is that, apart from providing estimated values for any spatial location, it also provides a variance function that indicates the accuracy of estimation for any point in the map. In practice this means that not only it is possible to generate a map of any variable across the field, but also to generate a map of the variance of the same variable from any given set of samples.
From a robotic exploration point of view, using such a variance map as a reward function for the exploration process will drive the robot towards the areas of the field where the information is less precise, improving overall model quality as shown previously for robotic mapping over time~\cite{santos2016lifelong}.
This paper makes the following contributions:

\begin{itemize}
\item a 3D representation of soil compaction based on observations at different spatial locations and depths;
\item use of a Kriging variance map as a reward function for robotic exploration, enabling adaptation of the robot's exploration strategy on-the-fly based on the current state of the map, resulting in more efficient data collection and better soil models;
\item an experimental comparison of common robotic exploration strategies for soil mapping using pre-recorded soil compaction data; and
\item a real-world dataset comprising a grid of sampled soil compaction measurements from a field environment, together with a reference model obtained by a soil scientist, which we will make available to the scientific community for benchmarking and result comparison following acceptance of the paper for publication.

\end{itemize}

The remainder of the paper is structured as follows: Section~\ref{sec:related} presents related work in soil surveying and robotic exploration, followed by Section~\ref{sec:method} which details our approach to adaptive soil sampling using a mobile robot. The experimental framework is presented in Section~\ref{sec:experimental}, followed by results and their analysis in Section~\ref{sec:results}, and final conclusions in Section~\ref{sec:conclusions}.



%% file: src/related.tex
\section{RELATED WORK} \label{sec:related}

Traditionally, soil condition maps are created from data collected manually at an arbitrary set of locations in the field, which are then used to create a map using geo-statistical tools such as Kriging. Kerry et al.~\cite{kerry2010sampling} show how Kriging variograms, an experimental method for defining spatial correlation between samples, have been deemed crucial for sampling planning in precision agriculture since the early 1990s.
They discuss the use of a priori or ancillary information to estimate a variogram and determine the spatial frequency of sampling depending on the range of the variogram. Other researchers estimate the variogram by a nested approach \cite{oliver1986combining}, where a set of samples is taken in a relatively small portion of the field and their variogram extrapolated to determine new sample locations.

Marchant \& Lark~\cite{marchant2007optimized} proposed an adaptive approach for optimizing reconnaissance surveys. They went into the field to be surveyed with a GPS and a laptop,  taking samples at pre-planned positions, following an initial plan. After each phase of data collection, the probability density function of the sampling density required for the main survey was calculated in a Bayesian framework. If the requirements were not met, the number and location of observations within further phases were selected to efficiently reduce the uncertainty of the estimate of the required sampling density. This Bayesian approach to adaptive reconnaissance leads to more efficient surveys than conventional approaches.

However, the effort required to survey a soil variable and simultaneously build and analyse the variance of the Kriging model of the soil, meant that soil scientists stopped short of planning the whole sampling procedure based on Kriging variance. Agricultural robots on the other hand, are already equipped with GPS and computing systems, and are able to create and update models of their operational environments through robotic exploration.

Robotic exploration approaches are usually aimed at creating a model of the robot's operational environment. 
%
Most previous research was aimed at developing efficient methods for creating environment representations that the robot could use to navigate and localise itself. 
These methods typically try to optimise the operation time against the completeness and accuracy of the world model. 
A common approach is to plan trajectories that completely cover the area assuming some prior knowledge of the environment~\cite{Bochtis2017}. 
If this is not possible, well-known exploration techniques drive the robot towards unmapped areas of the environment. 
For example, greedy approaches such as \cite{koenig2001greedy} drive the robot towards the closest location where new information can be gained. 
In frontier-based exploration~\cite{holz2010evaluating}, the robot plans its path in order to visit the boundary between the known and unknown parts of the environment, ensuring the completeness of the model but sacrificing time efficiency and model quality.
Information driven `next-best-view' methods use reward functions to predict the utility of a location~\cite{gonzalez2002navigation}.

Other  exploration methods aim to build models for other purposes besides robot navigation.
\cite{fentanes2011new} presents a next-best-view approach to build a 3D model of an outdoor environment, while maximising model quality and optimising the robot's trajectory for a search and rescue scenario.
Blaer and Allen~\cite{blaer2009view} proposed next-best-view methods to reconstruct 3D models of buildings. This solution benefited from a map of the environment, which was used to find the combination of viewpoints that requires the lowest number of scans to entirely cover the area. A second view planning step is used to cover all unpredicted occlusions in the model with as few scans as possible. This replanning stage can be considered as an example of adaptive sampling, which has received considerable attention in recent literature.

Adaptive sampling techniques can be considered as exploration methods that use a priori information from the environment to generate a plan that can be modified depending on the observations made. This is particularly useful when modelling physical phenomena that have an unknown spatial distribution. 
Dunbabin and Marques~\cite{dunbabin2012robots} show examples of adaptive sampling approaches for robotic environmental monitoring applications. Szczytowski at al.~\cite{szczytowski2010asample} present a method that tries to create a mobile network of sensors, where their spatial distribution is optimised to guarantee that every region of the environment is sampled uniformly in terms of information gain, using a Voronoi-based solution. 
Hombal et al.~\cite{hombal2010multiscale} present an adaptive sampling method where different strategies are used to optimise mission constraints such as time, speed and sampling uniformity. They further present an interesting comparison of the results obtained with different strategies when trying to optimise different variables. Finally Thompson and Wettergreen~\cite{thompson2008intelligent} use a priori information obtained by remote sensing techniques combined with locally measured data through Gaussian process models to predict where to maximise information discovery by the robot, adapting its trajectory accordingly.

This paper presents an extended comparison between different exploration strategies in a soil surveying application. 
Our exploration methodology extends previous research by applying adaptive sampling techniques to build soil condition maps with an autonomous mobile robot, resulting in more efficient data collection and better quality soil models.

%% file: src/method.tex
\section{METHODOLOGY} \label{sec:method}

In our scenario, a mobile robot equipped with a penetrometer is used to generate a 3D soil compaction map of an agricultural field. Typically, soil condition maps are created from data samples collected in the field, which are then extrapolated over the working area using geostatistical techniques. These tools quantify the spatial autocorrelation among samples and utilise this to make a prediction for a location in the field and the variance of this prediction. Usually these methods are known as Kriging or Gaussian process regression. 

They are considered interpolation methods that predict values without bias.
There are different types of Kriging techniques, such as Ordinary Kriging (OK), Universal Kriging (UK), Indicator Kriging and Co-Kriging, among others,  their main difference being the trend used to estimate the model variation across the working area. The most commonly used method is OK, which is the one used in the work presented.

\subsection{Ordinary Kriging (OK)}

Ordinary Kriging provides an estimate $\hat{Z}(\mathbf{x}_0)$ for a variable $Z$ at unknown location $\mathbf{x}_0$ whilst assuming a constant unknown mean over its neighbourhood. The estimate is a weighted linear combination of the $n$ available (i.e. observed) values $z_i=Z(\mathbf{x}_i)$ at a set of locations $\mathbf{x}_i$ which minimises the variance of errors:
\begin{equation} \label{eq:estimator_}
\hat{Z}(\mathbf{x}_0)=\sum_{i=1}^{n}w_iz_i, \quad i=1,\dots,n,
\end{equation}
where $\sum_{i=1}^{n}w_i=1$ to assure unbiased estimates. The weights $\mathbf{w}=[w_1,\dots,w_n]^T$ depend solely on the distance between the locations $\mathbf{x}_i$ and are independent of the actual values $Z(\mathbf{x}_i)$.
To deduce weights which the available samples will have on the estimation at location $\mathbf{x}_0$ the following system of equations needs to be solved:
\begin{equation} \label{eq:OK_}
	\begin{multlined}
      \begin{bmatrix}
      \mathbf{w} \\ 
      \lambda
      \end{bmatrix}
      =
      \begin{bmatrix}
       \mathbf{C}_{ij} & \mathbf{1} \\ 
       \mathbf{1}^T & 0
      \end{bmatrix}^{-1}
      \begin{bmatrix}
      \mathbf{C}_{i0}\\ 
      1
      \end{bmatrix}
	\end{multlined},
\end{equation}
where $\mathbf{C}_{ij}=Cov[Z(\mathbf{x}_i),Z(\mathbf{x}_j)]$ is the covariance of the observed values, $\mathbf{C}_{i0}=Cov[Z(\mathbf{x}_i),Z(\mathbf{x}_0)]$ is the covariance at the prediction location $\mathbf{x}_0$ and $\lambda$ is a Lagrange factor which ensures the optimal solution.

Once this system is solved, the estimated values at location $\mathbf{x}_0$ can be found using Eq.~\ref{eq:estimator_}, and the associated variance of the prediction $\sigma^2$ can be calculated as follows:
\begin{equation} \label{eq:variance}
\sigma^2(\mathbf{x}_0) = \sum_{i=1}^{n}w_i\mathbf{C}_{i0}. 
\end{equation}

Our application considers a soil compaction variable, which tends to gradually increase its value with soil depth. Since OK assumes a constant mean in the close neighbourhood of the sampled location, we cannot use this Kriging method directly to model the soil compaction at different depths. Therefore, we discretise the soil depth into a set of $m$ layers of a specific interval and create a separate Kriging model for each layer together with its variance. The general Kriging variance KV is an average of the mean Kriging variance KV\textsubscript{i} for each layer:
\begin{equation} \label{eq:kv}
KV=\frac{1}{m}\sum_{i=0}^m KV_i.
\end{equation}

\subsubsection{Semivariogram}

\begin{figure}[h]
	\centering
      \includegraphics[width=\columnwidth]{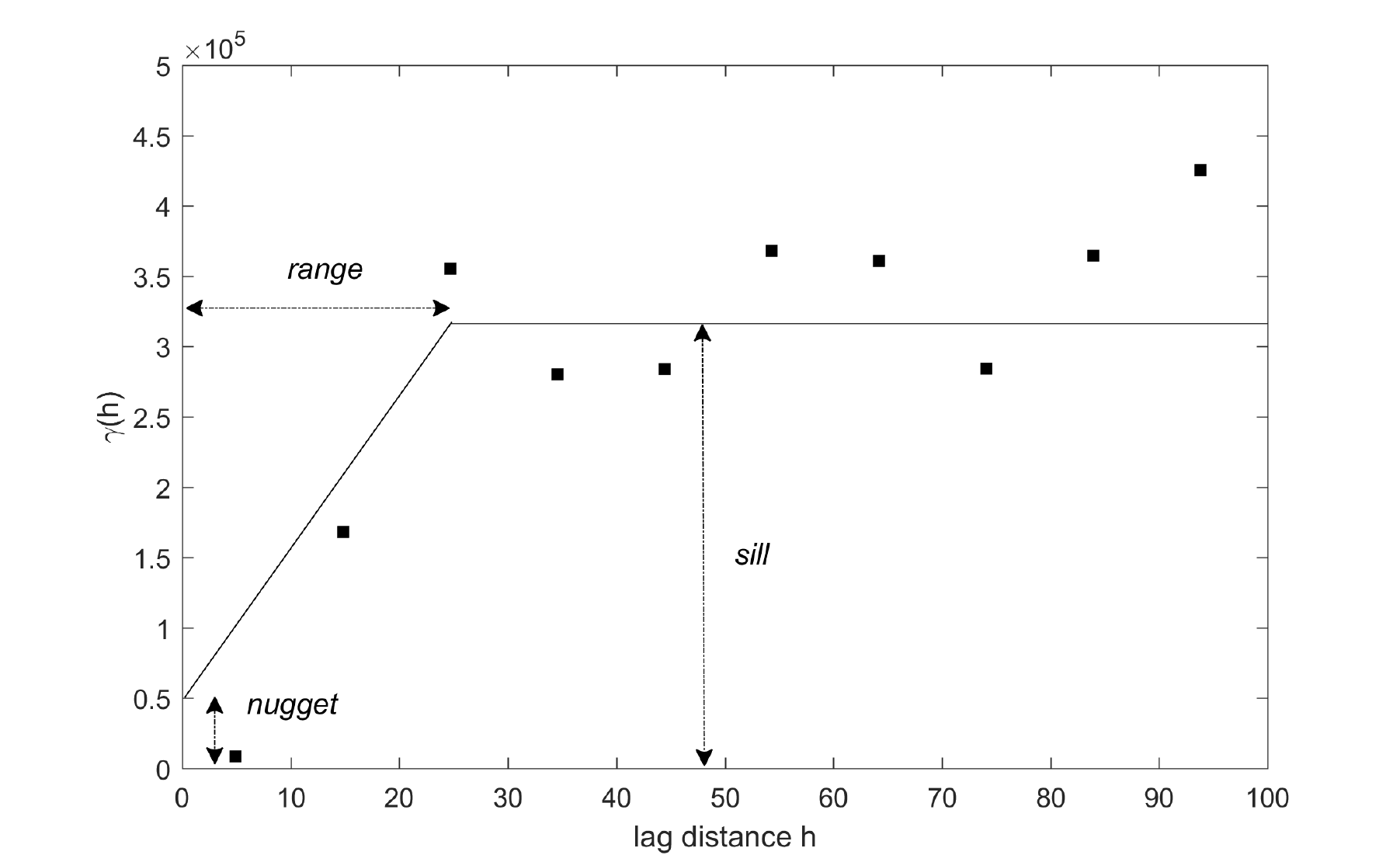}
      \caption{Semivariogram: \emph{nugget}, \emph{range} and \emph{sill} parameters ($p_0, p_1, p_2$) .\label{fig:semivariogram_models}}
\end{figure}

In practical applications, the theoretical covariances $\mathbf{C}$ in Kriging are replaced by semivariances $\mathbf{\Gamma}$ derived from experimental semivariograms which express the spatial correlation as a function of distance $\mathbf{h}$ between locations $\mathbf{x}_i$. The semivariograms $\mathbf{\gamma}(\mathbf{h})$ can take multiple forms but generally are characterised by three parameters (see Fig.~\ref{fig:semivariogram_models}): $p_0$ - \emph{nugget} is the semivariance at distance 0, $p_1$ - \emph{range} defines the distance that is spatially correlated and $p_2$ - \emph{sill} is the semivariance at distances beyond the \emph{range}. These parameters are obtained through a semivariogram fitting procedure. We use the following linear semivariogram model in our work: 


\begin{equation}\label{eq:linear_sv}
  \gamma(h)=
  \begin{cases}
    p_0 + p_2\frac{h}{p_1}, & 0 < h < p_1\\
    p_0 + p_2, & p_1 < h\\
    0, & h=0
  \end{cases}
\end{equation}

\subsection{Exploration Strategies}

Our proposal is to use the variance of the Kriging \emph{(KV)} process (see Eq.~\ref{eq:kv}) as a measurement of information gain. However, as reported in Section~\ref{sec:related}, there have been multiple proposals in the literature on how to efficiently explore an environment. In this work, we therefore compare several common strategies for using this information towards mapping soil compaction in the same environment. The tested strategies can be classified into three different categories: \emph{Area Coverage}, \emph{Next-Best-View} and \emph{Adaptive Sampling} methods.

\subsubsection{Area Coverage}

\begin{figure}[h]
	\centering
	\begin{subfigure}[b]{0.3\columnwidth}
      \includegraphics[width=\columnwidth]{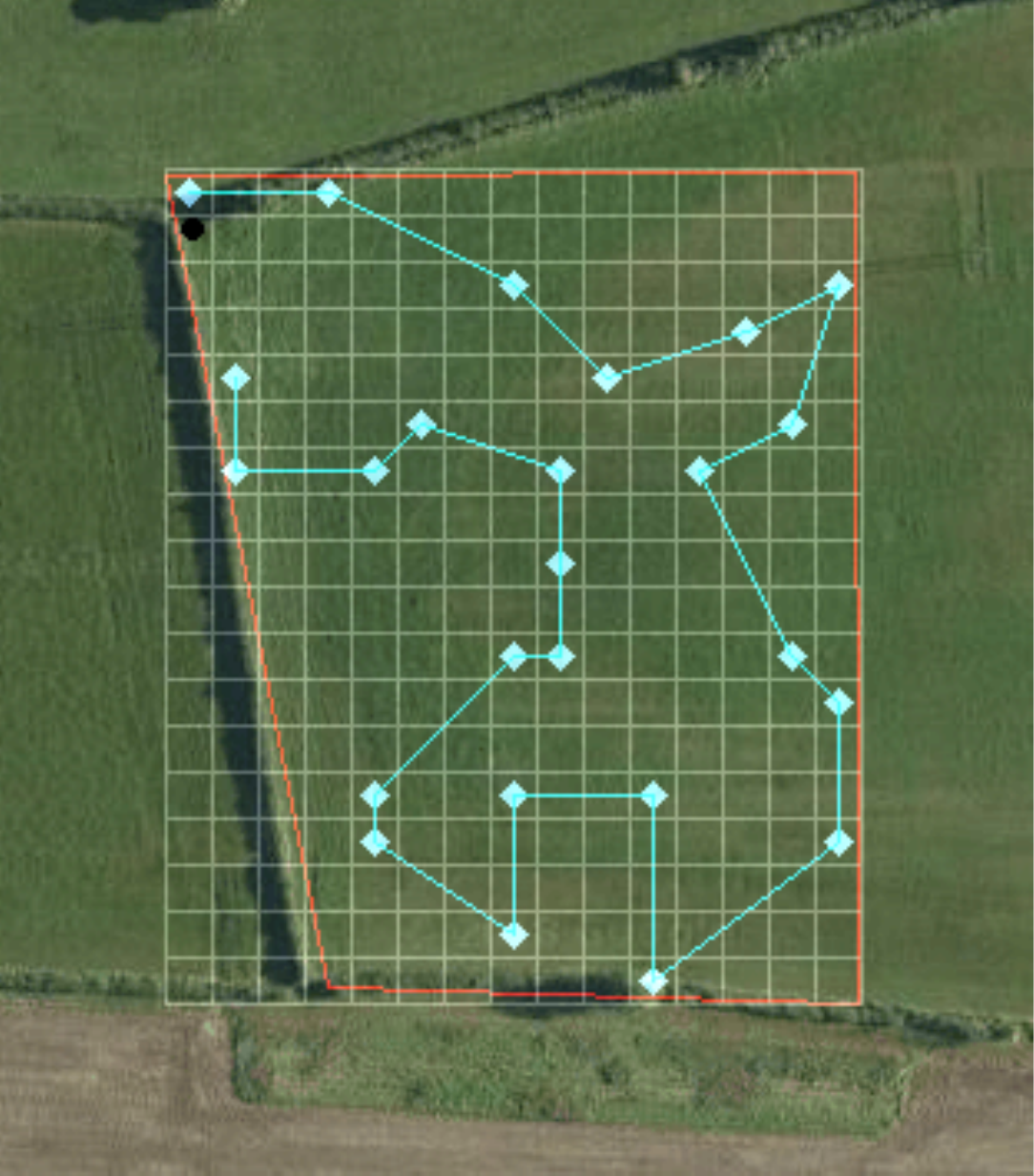}
      \caption{\label{fig:random}}
    \end{subfigure}
	~
    \begin{subfigure}[b]{0.3\columnwidth}
      \includegraphics[width=\columnwidth]{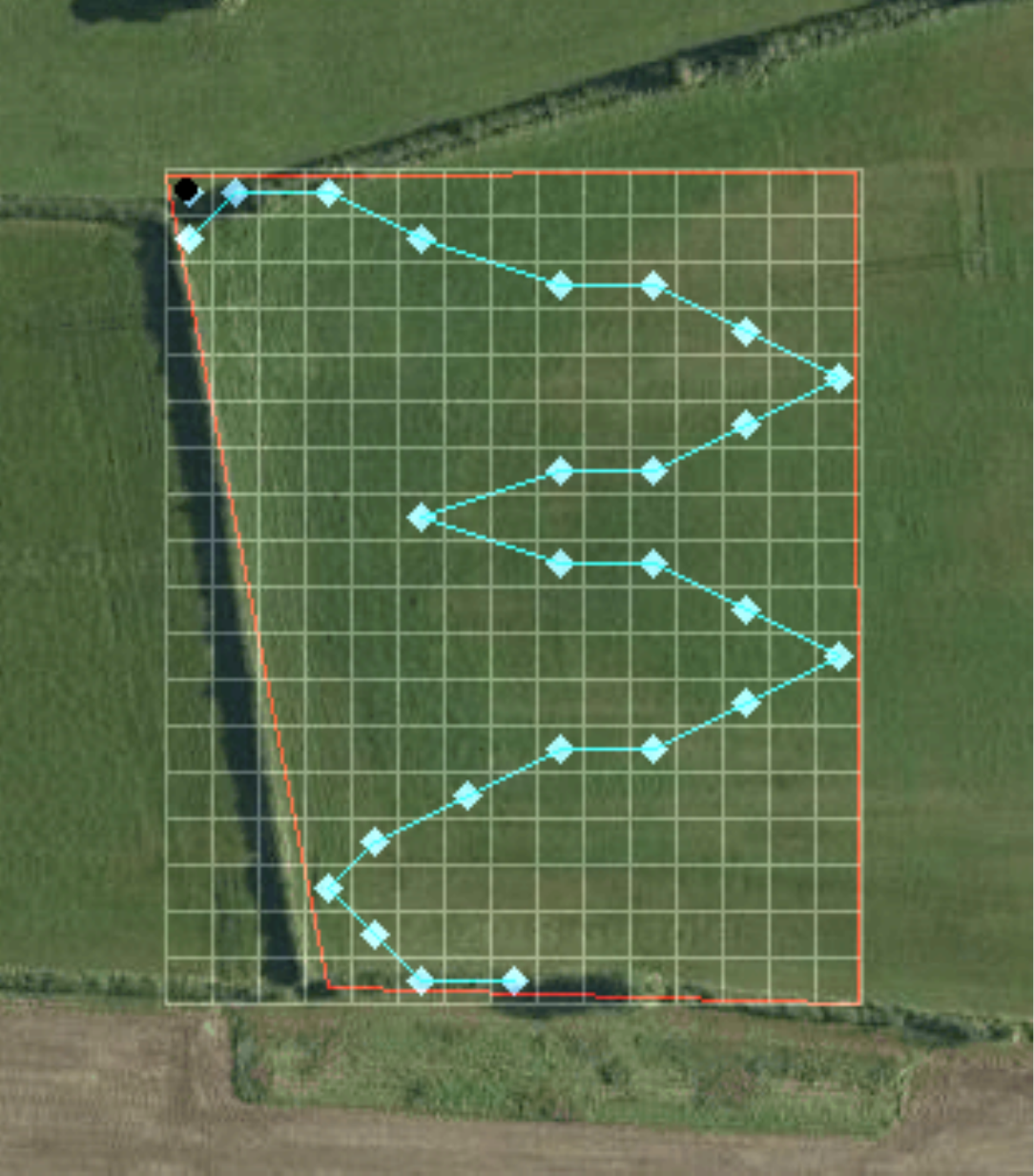}
      \caption{\label{fig:w_shape}}
    \end{subfigure}
    ~
    \begin{subfigure}[b]{0.3\columnwidth}
      \includegraphics[width=\columnwidth]{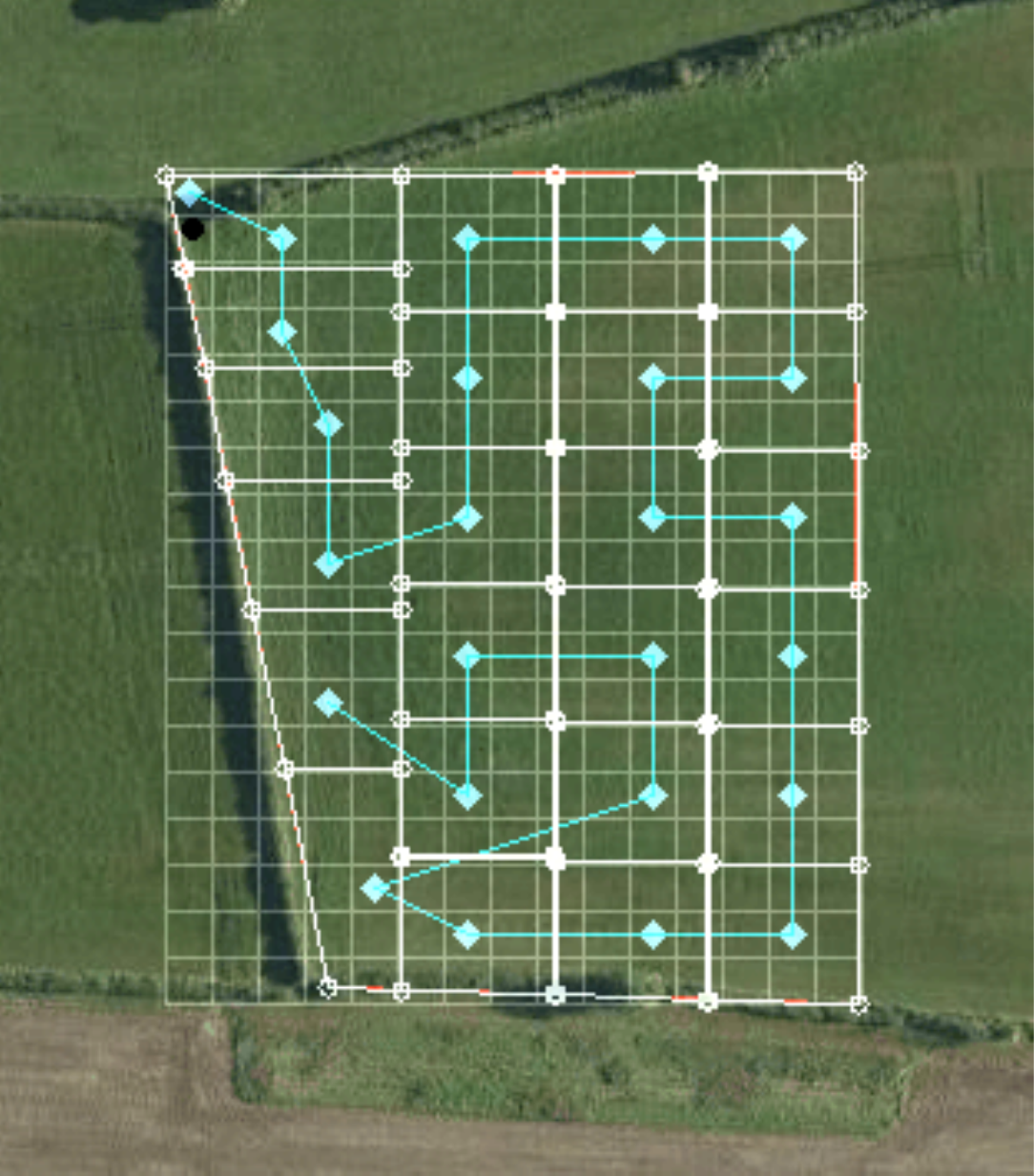}
      \caption{\label{fig:area_split}}
    \end{subfigure}
    \caption{(a) Random trajectory for area coverage, (b) \textbf{W} shape predetermined pattern and (c) Area splitting method for area coverage}
\end{figure}

These methods, unlike the other two categories, are not information based. This means that they \emph{do not} use the proposed KV for their planning. However, they are the most commonly used strategies for soil condition mapping, and provide a good  baseline comparison to see the effect of using KV as a reward function. In these methods, at the start of the exploration process the robot chooses its sampling locations in a specific manner and then plans the order in which to visit them using a \emph{tsp} (travelling sales-person) method to find the shortest path that connects them. The static coverage methods tested are the following:

\begin{itemize}
\item \emph{Random}: In this case, the robot goes to a set of randomly selected locations around the field, see Fig.~\ref{fig:random}.
\item \emph{Pre-determined pattern}: Sampling locations are allocated following a predetermined pattern, in our case a popular with soil scientists W-shaped path, across the field, see Fig.~\ref{fig:w_shape}.
\item \emph{Area splitting}: Here the field is divided into a specific number of polygons, each with roughly the same area, and a sampling location is added at the centre of these areas, see Fig. \ref{fig:area_split}
\end{itemize}

\subsubsection{Next-Best-View (NBV)}

These methods update the environment model every time a new sample is acquired and then choose a new location depending on the distribution of the KV across the field. Location selection is done using one of the following strategies:

\begin{itemize}
\item \emph{Greedy}: The next sampling point is the point with the highest KV in the set of candidate locations.
\item \emph{Monte Carlo}: a set of candidate sampling locations is generated each time, and each candidate location is allocated a weight depending on its KV. The next sampling location is selected randomly but in a way that guarantees that probabilities are distributed according to the weight of each candidate.
\end{itemize}

\subsubsection{Adaptive Sampling}

In this category, strategies generate an initial plan that is modified depending on the KV after each model update. In this case, the robot will plan a sampling regime based on the static coverage strategies.
However, every new sample taken triggers a model update, which is then used to recalculate the sampling regime.
This recalculation includes adding or removing sampling points based on a \emph{NBV} strategy and their KV, and replanning a new route through the new set of sampling points using a Travelling Salesman Problem (TSP) algorithm. For this purpose we verify the following combinations of strategies:

\begin{itemize}
\item \emph{Adaptive + Greedy}: An initial path is generated by either random or area sampling and new targets are added whenever there is a cell with a higher KV than any of the ones in the current plan, targets are removed when their KV is less than $2 \sigma$ of the distribution of the KV of the targets in the current plan. Every time a plan is modified, a new trajectory is calculated using a TSP solver.   
\item \emph{Adaptive + Monte Carlo}: As in the previous method an initial path is generated, however in this case new targets are added by drawing a new waypoint from a set of candidates weighted by their KV, the target with the lowest KV in the current plan is removed to keep the route with equal number of waypoints.
\end{itemize}


%% file: src/experimental.tex
\section{EXPERIMENTAL FRAMEWORK} \label{sec:experimental}

\subsection{Thorvald and Penetrometer}

Our experimental set-up consists of an autonomous mobile robot Thorvald equipped with a custom made automatic penetrometer device for measuring soil compaction (see Fig.\ref{fig:thorvald_infield}). Thorvald \cite{grmistad17thorvald} is an autonomous, general purpose, light-weight outdoor mobile platform designed for agricultural applications. Its modular design allows for easy re-configuration for a specific application and environment. 
For example, the width of the platform can be adjusted for different applications, crop row widths, etc.
In our work, we used a standard configuration consisting of four-wheel driving and steering, and 1.5 m width to ensure stable operation of the automated penetrometer. The robot batteries provide roughly 8 hours of autonomy without the payload. The robot is controlled through an in-built PC running Linux OS and Robot Operating System (ROS). The platform is equipped with a GNSS sensor, which enables robot localisation and geo-tagging of the collected data samples. The navigation component uses a graph-based representation, allowing the robot to move along a pre-defined path.

The automated penetrometer consists of a steel rod with a coned tip driven into the soil by a linear actuator mounted vertically on a two-axis Cartesian gantry, enabling convenient horizontal local movement of the penetrometer without the necessity of moving the entire robot. The probing rod is equipped with a force sensor (iLoad Pro Digital USB Load Cell by Loadstar), providing continuous force readings during operation of the device. The vertical actuator provides additional feedback, allowing excessive pushing forces to be detected when the rod hits a hard surface. In such situations, the rod is withdrawn from the soil, the gantry moves the penetrometer to an alternative nearby location and another sampling attempt is made. The penetrometer was designed to exert a maximum pushing force of 600 N whilst the range of the force sensor is up to 1100 N. The actuators are moved by industrial motor controllers which communicate with the robot software infrastructure through a ROS driver.

A compaction measurement at a single location consists of a series of force readings whilst the rod is being pushed into soil with a constant speed of 2.5 cm/s (see Fig.\ref{fig:thorvald_infield}). In our case, the rod can reach soil depths of 50 cm and each sample consists of 300 force readings. The entire sampling procedure including probing, withdrawal of the probe and safety delays at a single location takes around 50 s. The entire system's bandwidth, taking into account the sampling and robot movement is therefore around 30-60 samples/h.

\subsection{Data Collection}

\begin{figure}[h]
	\centering
	\begin{subfigure}[b]{0.3\columnwidth}
      \includegraphics[width=\columnwidth]{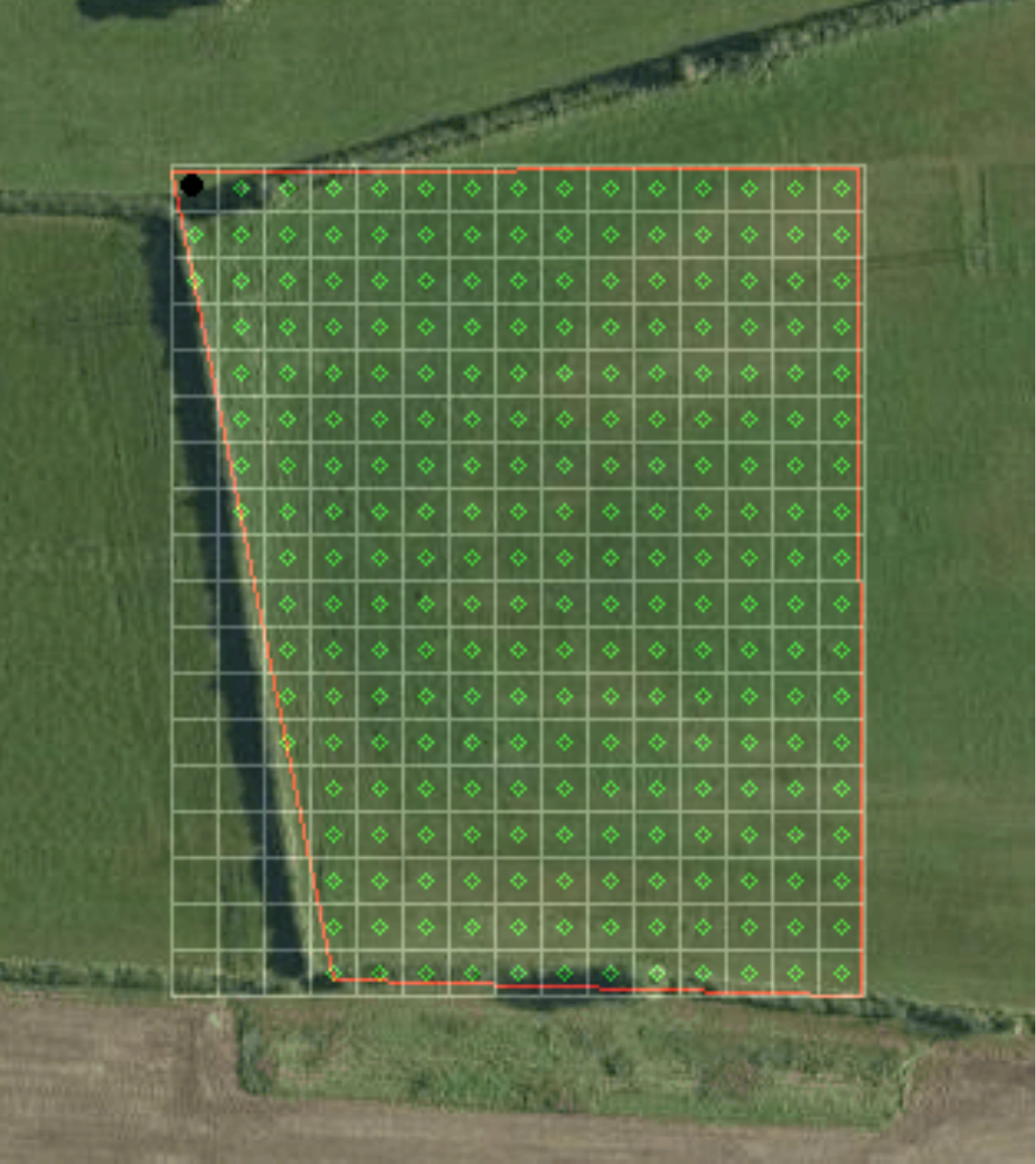}
      \caption{\label{fig:waypoints}}
    \end{subfigure}
	~
    \begin{subfigure}[b]{0.3\columnwidth}
      \includegraphics[width=\columnwidth]{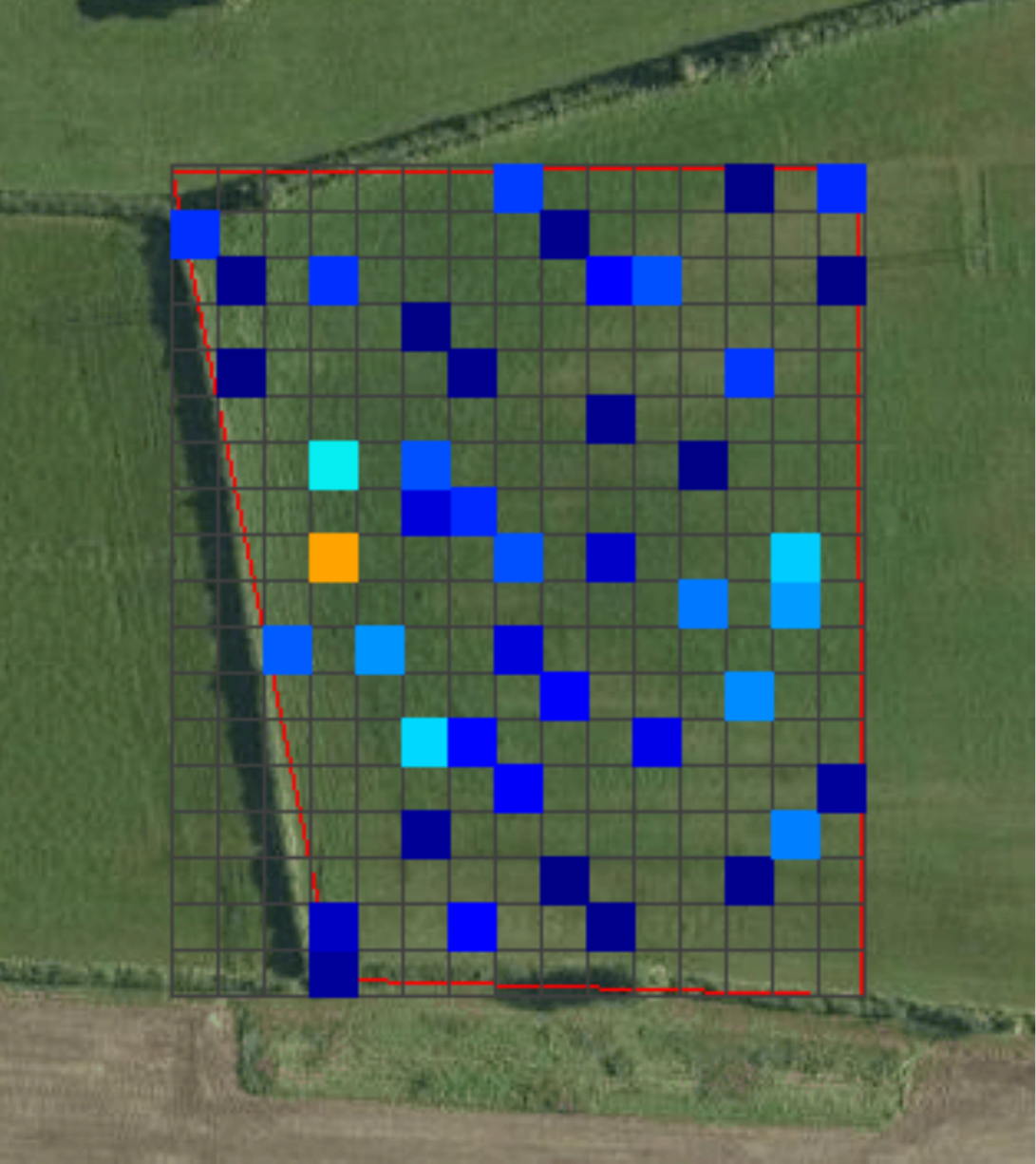}
      \caption{\label{fig:manual}}
    \end{subfigure}
    ~
    \begin{subfigure}[b]{0.3\columnwidth}
      \includegraphics[width=\columnwidth]{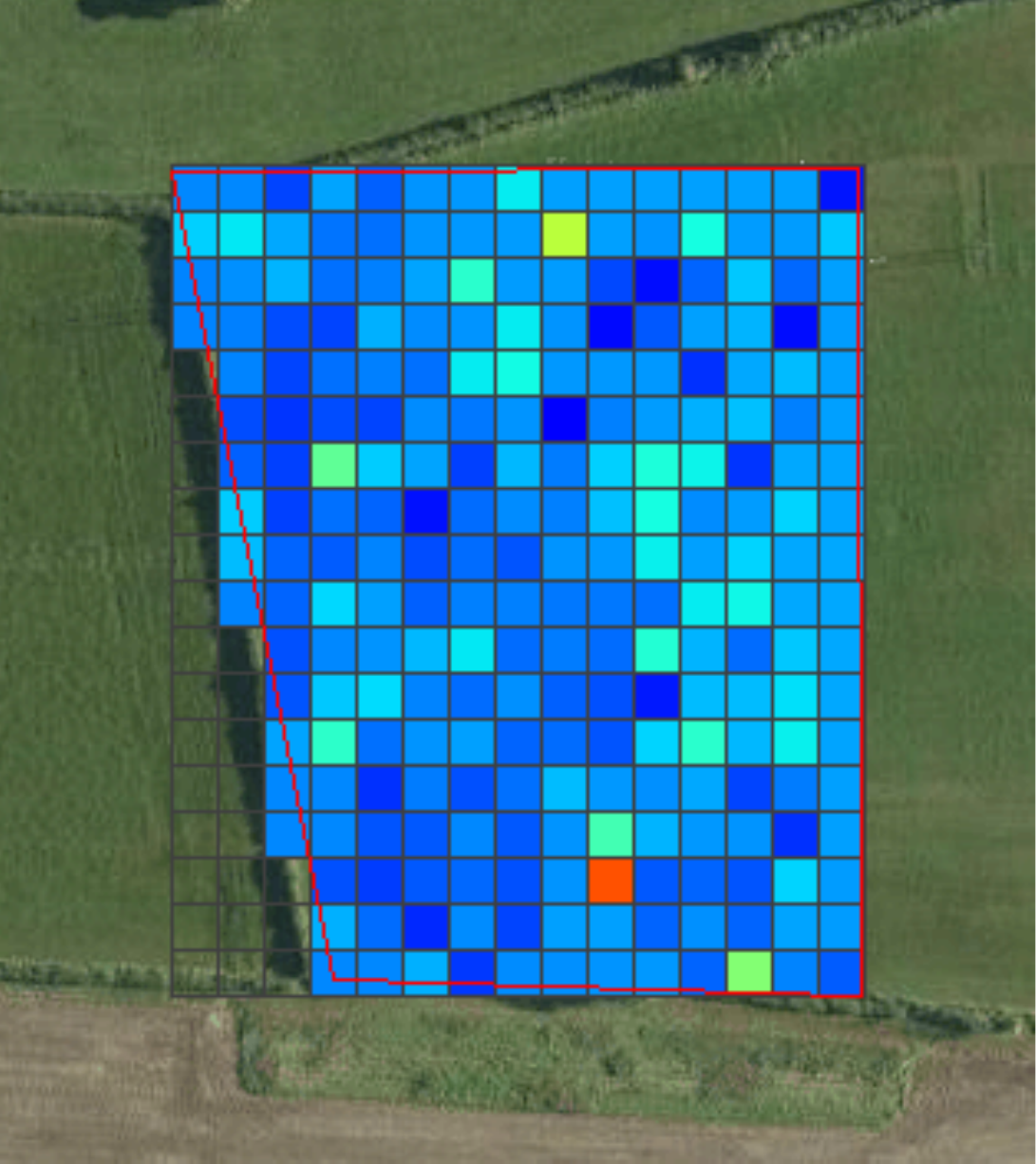}
      \caption{\label{fig:high_density}}
    \end{subfigure}
    \caption{Experimental field at the Riseholme campus, Lincoln, UK: (a) A regular 10$\times$10 m sampling grid with green circles indicating navigation way-point locations in a 2.33 ha area; (b) Approximate locations of manually collected compaction samples (in colour); (c) Locations of automatically collected compaction samples (in colour).}
\end{figure}

Data was collected from a field site at the University of Lincoln's Riseholme Campus, Lincolnshire, UK (OS grid reference 498456 374863). The field was in permanent pasture (cattle and sheep) at the time of sampling, and sits on a  clay soil over limestone (\textit{Elmton 1} association \cite{hodge1984soils}, with slight slope falling towards the west. We consider a roughly rectangular area of 2.33 ha for our experiments (see Fig.~\ref{fig:waypoints}).

An initial dataset was collected manually with the purpose of calibrating the autonomous mobile sampling and Kriging model differences. 49 manual readings of soil compaction were taken with a hand-held digital penetrometer (FieldScout SC900) on a loose grid system across the field, with approximate spacings of 25m between sampling locations (see Fig.~\ref{fig:manual}). Each manual sample  requires around 50 s to complete which includes manual probing and GPS tagging~-- this figure is very similar to the sampling time required by our automated penetrometer.

For the high density data collection, the field was divided into 10$\times$10 m cells with a way-point situated in the centre of each cell. At each way-point, the robot took a penetrometer reading together with a GPS location which resulted in 286 geo-tagged samples (see Fig.~\ref{fig:high_density}). 


\subsection{Surrogate Model}\label{sec:surrogate}

Benchmarking robot decisions is a difficult task. 
Different decisions can lead to the observation of phenomena in different ways. 
This means that to make an accurate comparison, a `ground truth' model of the phenomenon investigated is needed that is better than the ones obtained by the methods being compared, in order to provide a fair baseline for comparisons. Also, 
comparing models generated directly from data captured by the robot has an additional complication that field conditions might have changed from test to test, making such comparisons unreliable.
Therefore, in this work,
the high density dataset was used to create a dense `surrogate' model of soil compaction, which we then use in off-line `simulations' to compare different exploration strategies and understand their overall performance. Simulations using a surrogate model are a useful tool to compare exploration methods~\cite{fentanes2011algorithm,Santos4DSim}, providing the `ground truth' for the resulting soil compaction models.

%% file: src/results.tex
\section{RESULTS} \label{sec:results}

\begin{figure*}[t]
	\includegraphics[width=\textwidth]{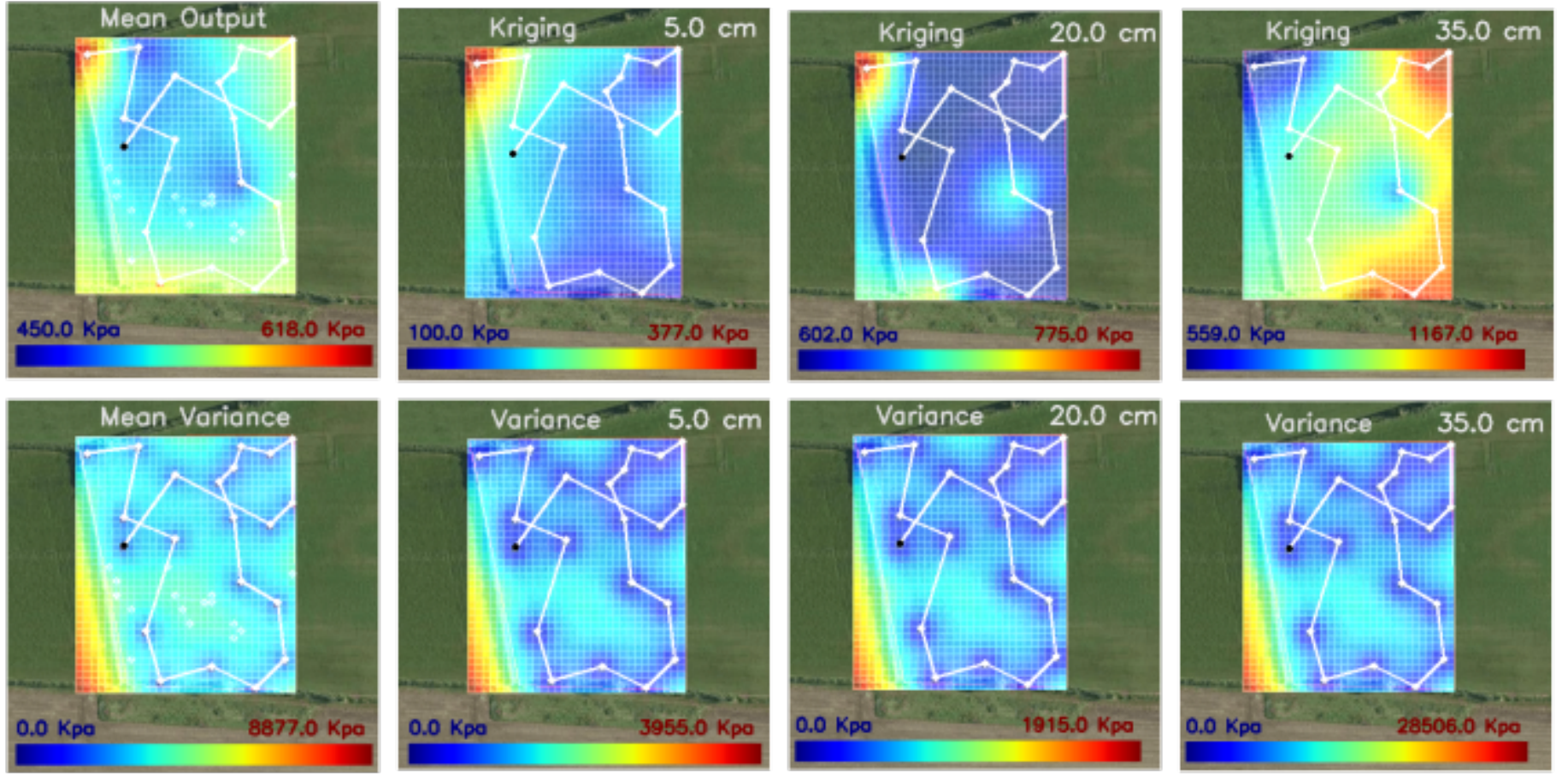}
    \caption{Soil compaction maps generated with an adaptive sampling strategy, white line indicates the path followed by the robot\label{fig:compaction_maps}}
\end{figure*}

The results presented in this section were obtained using simulated runs over the surrogate model presented in Section~\ref{sec:surrogate}. For our comparisons, we divided the field into a $5\times5$ m grid, resulting in 936 reachable cells over which the exploration algorithms can be executed. We divided the soil depth into $m=8$ layers of 5 cm each, for which a separate Kriging model together with its variance is calculated. Figure \ref{fig:compaction_maps} shows soil compaction maps and their variance for different layers and the global mean Kriging variance used for exploration.

To compare any two resulting soil compaction models $G$ and $C$ we propose to use the Mean Square Error (MSE):
\begin{equation} \label{eq:rmse}
MSE=\frac{1}{N}\sum_{i=1}^{n}\left ( G_i - C_i\right )^2,
\end{equation}
but also its square rooted variant RMSE to keep unit consistency and clarity of presentation.

\subsection{Manual vs. Automated Sampling Soil Models}

\begin{figure}[h]
\centering
	\includegraphics[width=0.6\columnwidth]{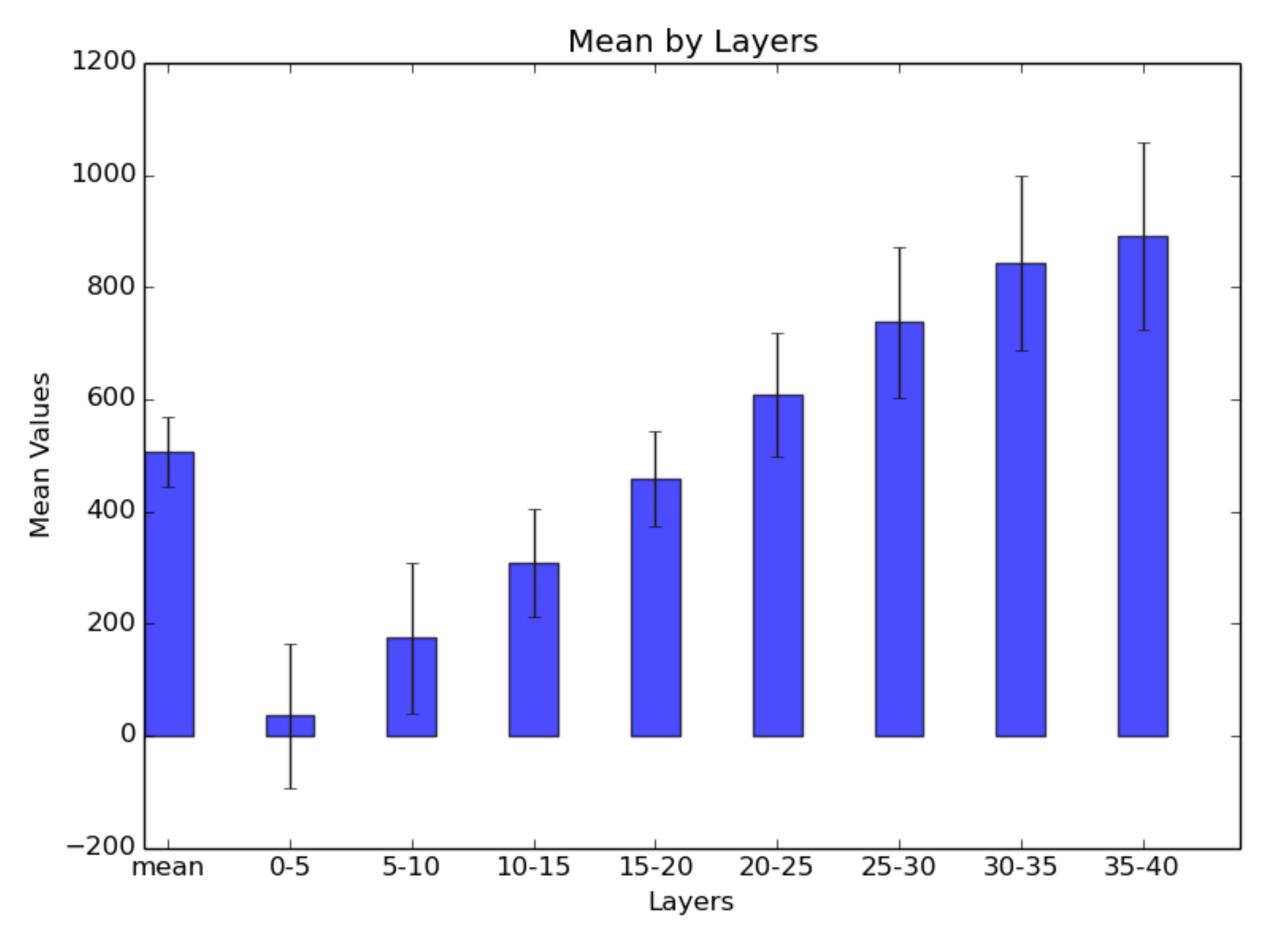}
    \caption{Mean compaction values for the surrogate model. The error bars show the RMSE between surrogate and manual models.\label{fig:comparison}}
\end{figure}

To ensure the consistency and validity of our surrogate soil compaction model, we compared it to the model created from the manually collected samples. It is important to point out that the manually created model cannot be treated as definite ground truth since its creation is subject to numerous biases related to the quality of the instrument and sampling procedure. Figure~\ref{fig:comparison} shows the mean values of the high density model together with model error (RMSE) in the error bars. The variability is similar throughout the soil layers,  generally indicating a good alignment between the models, even though the top soil (0-20 cm) is characterised by larger deviations. These differences might be due to multiple factors such as probe engagement with the soil in automated and manual sampling,the time difference (4 months) between acquisition of both datasets, and the top soil being more susceptible to short- and medium-term changes in compaction. Table~\ref{tab:RMSE_Mean} shows that the normalised RMSE between both models in the deeper layers is the same, indicating a small bias which might be caused by a calibration difference between the two methods. 

\begin{table}
\centering
\caption{Normalised RMSE between the surrogate and manual models for each layer.}
\label{tab:RMSE_Mean}
\begin{tabular}{cc}
Layer & RMSE/Mean  \\\midrule
0-5   & 3.61       \\
5-10  & 0.77       \\
10-15 & 0.31       \\
15-20 & 0.18       \\
20-25 & 0.18       \\
30-35 & 0.18       \\
35-40 & 0.18       
\end{tabular}
\end{table}

\subsection{Kriging Variance for Exploration}

Our proposal is based on the assumption that reducing Kriging variance \emph{(KV)} will lead to a better model. In other words, we assume that KV is correlated to the model error. To verify this hypothesis, we simulated different exploration strategies and compared the model error (MSE) and Kriging variance (KV). Both values were calculated for each cell of the model and used to generate two vectors for each sample. These vectors were then used to obtain the correlation coefficient between the two variables. 

\begin{table}[h]
\caption{Correlation between Kriging variance (KV) and model error (MSE).}
\label{tab:Correlation}
\centering
\begin{tabular}{@{}lcc@{}}
Strategy	&  \#Runs & Corr. coeff.\\ \midrule
Greedy		&		5		& 0.79 \\
Random     	& 		5	    & 0.84 \\
Area Split 	& 		1 		& 0.98 \\
AS + Greedy &		5		& 0.96 \\
\end{tabular}%
\end{table}

\begin{figure}[h]
	\centering
	\begin{subfigure}[b]{0.44\columnwidth}
      \includegraphics[width=\columnwidth]{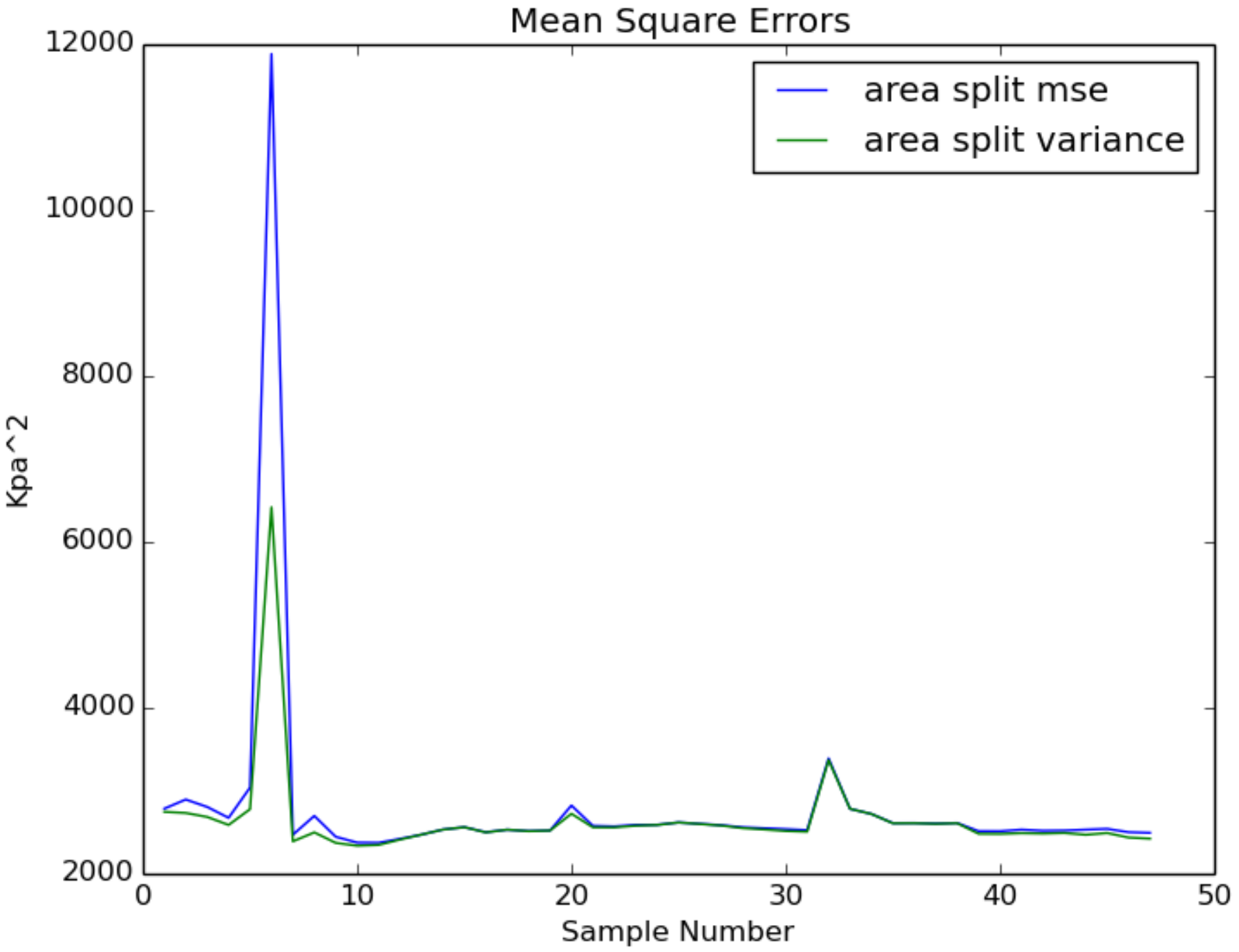}
      \caption{\label{fig:as_var_mse}}
    \end{subfigure}
	~
    \begin{subfigure}[b]{0.44\columnwidth}
      \includegraphics[width=\columnwidth]{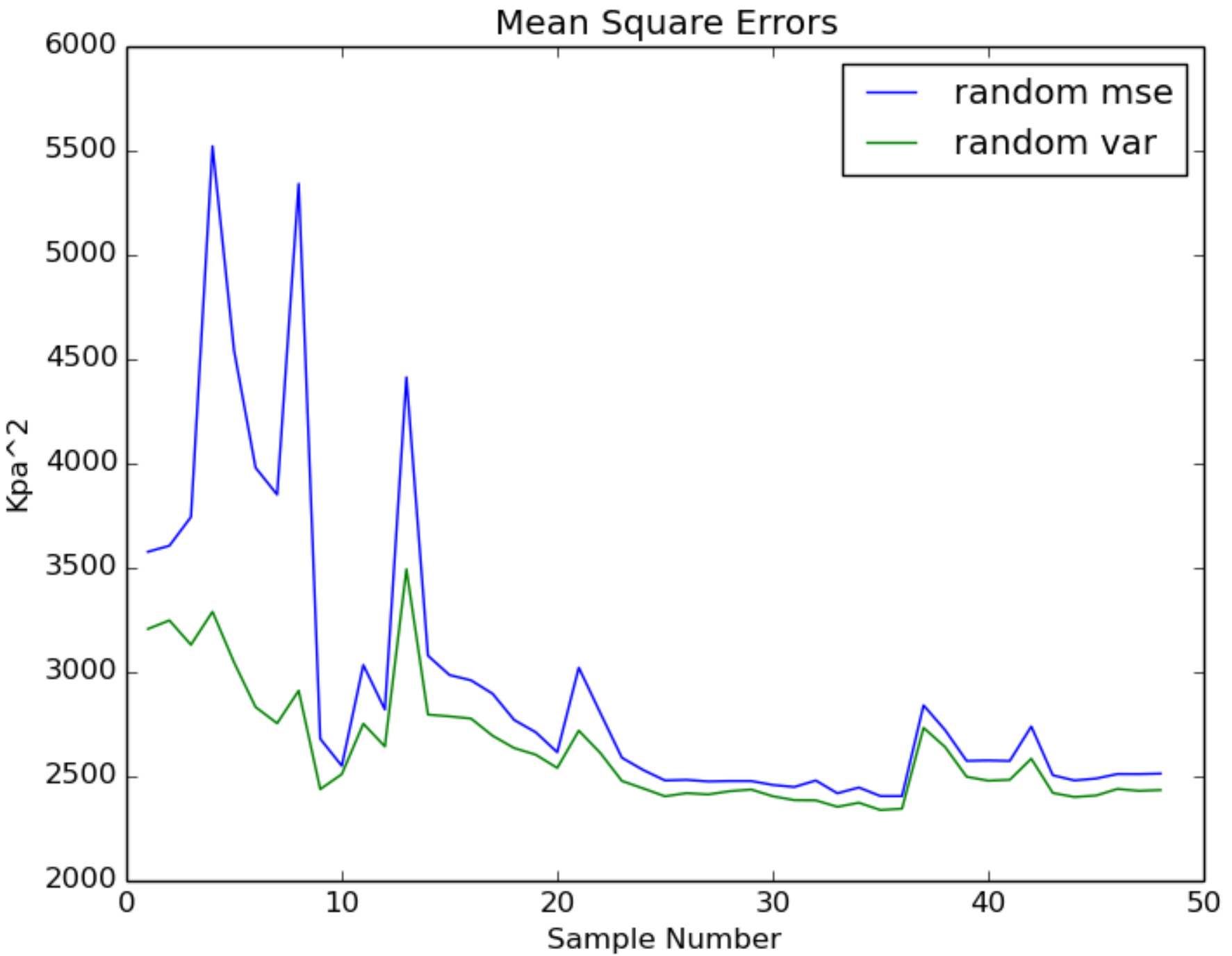}
      \caption{\label{fig:rand_var_mse}}
    \end{subfigure}
    \\
    \begin{subfigure}[b]{0.44\columnwidth}
      \includegraphics[width=\columnwidth]{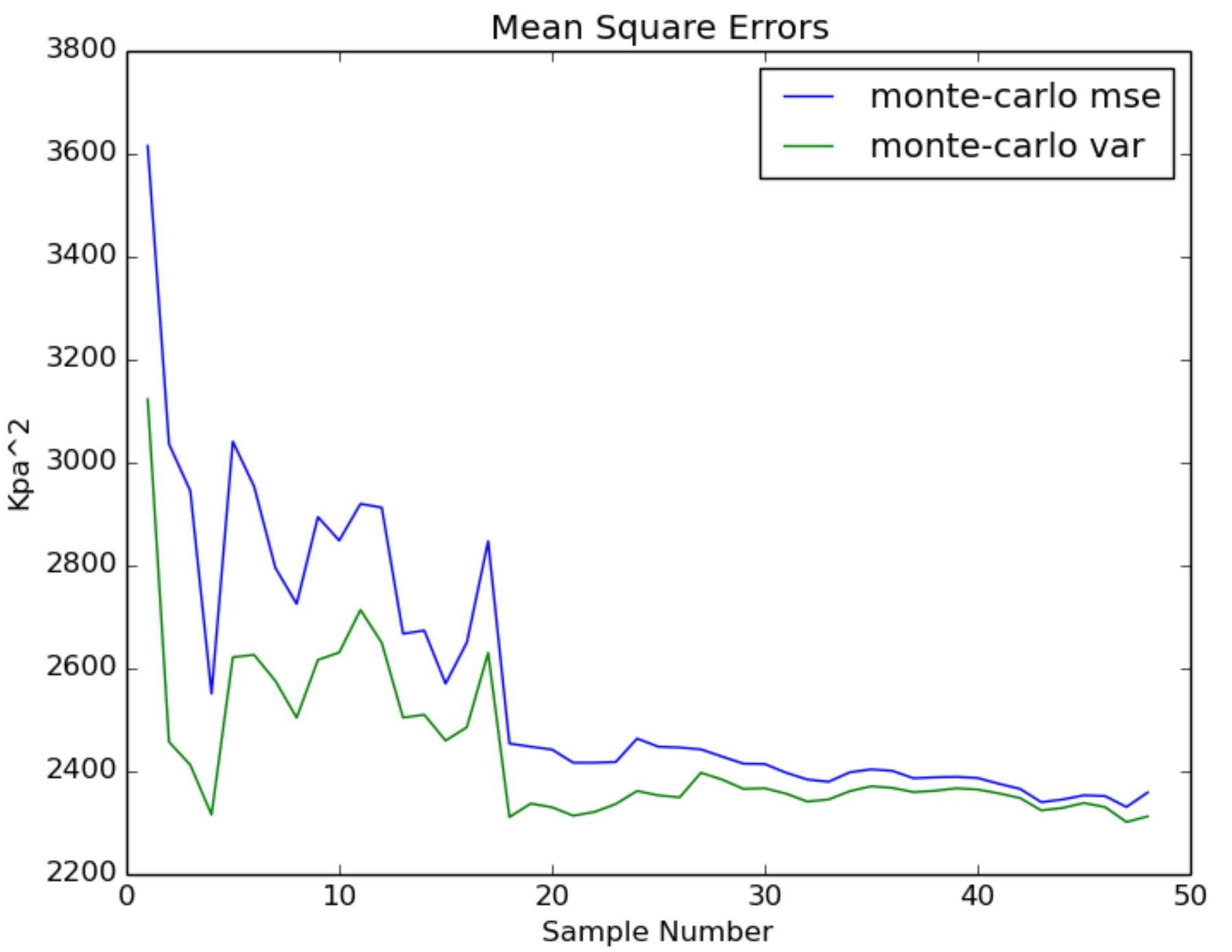}
      \caption{\label{fig:as_var_mse}}
    \end{subfigure}
	~
    \begin{subfigure}[b]{0.44\columnwidth}
      \includegraphics[width=\columnwidth]{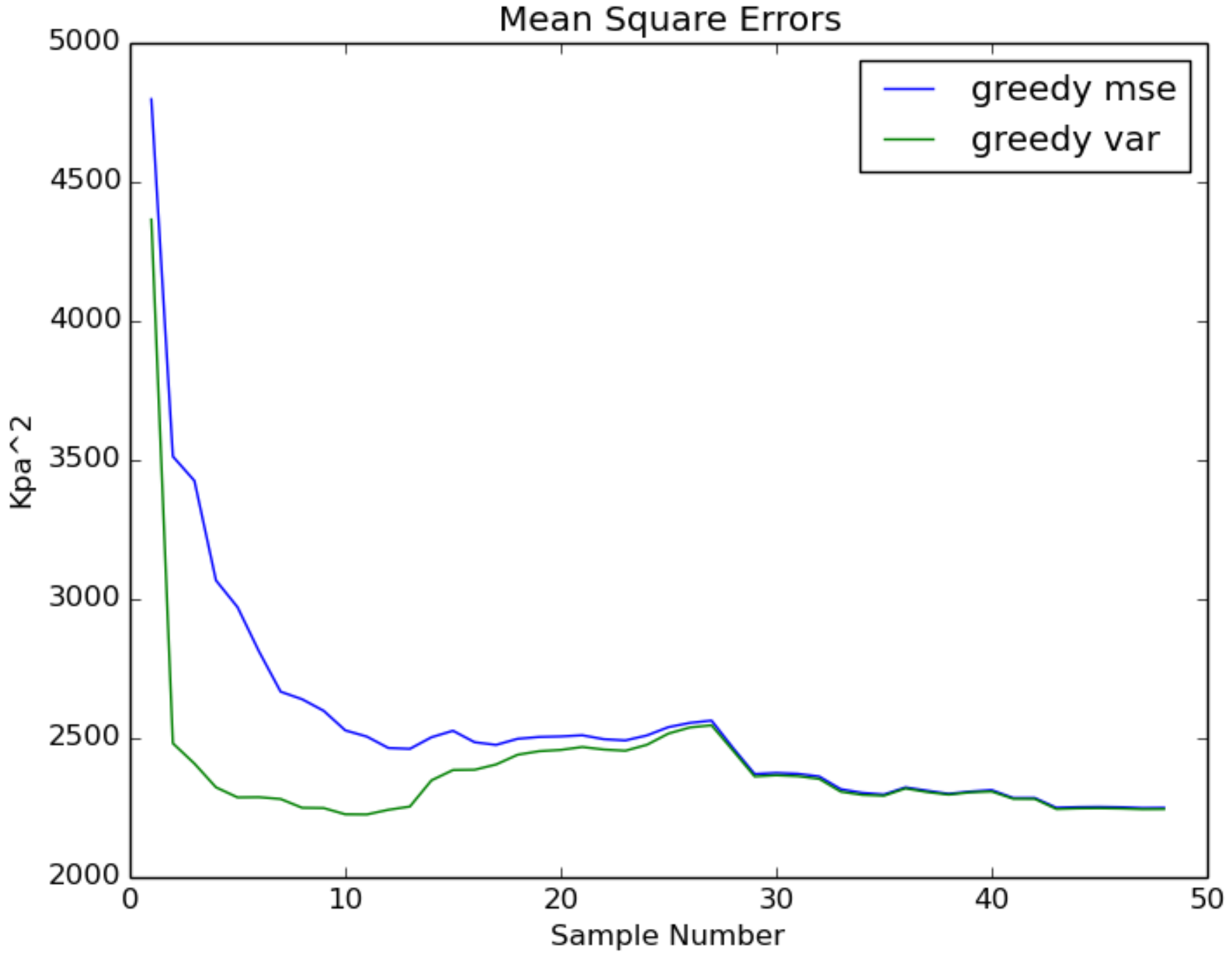}
      \caption{\label{fig:rand_var_mse}}
    \end{subfigure}
    \caption{Model error (MSE) and Kriging variance (KV) over time different exploration strategies: (a) Area Splitting; (b) Random; (c) Monte-Carlo; (d) Greedy. \label{fig:var_mse}}
\end{figure}

Table~\ref{tab:Correlation} shows the correlation coefficient between MSE and KV for multiple exploration strategies, indicating a strong correlation between these variables. Figure~\ref{fig:var_mse} further illustrates the consistency of this correlation over time for different exploration strategies and simultaneous convergence of both values as more information is incorporated into the models.


\subsection{Exploration Strategy Performance}

In the previous section, we showed how reducing the Kriging variance will lead to higher quality models. In practice, other factors such as time or energy needed to create the models must also be taken into account when evaluating exploration strategies. Table~\ref{tab:RMSE_samples} shows the dependence of the model error on the number of samples taken for different strategies.


\begin{table}
\centering
\caption{Model error (RMSE, kPa) per number of samples.}
\label{tab:RMSE_samples}
\begin{tabular}{lcccc}
                  & \multicolumn{4}{c}{\#Samples}                       \\ 
\cline{2-2}\cline{3-3}\cline{4-4}\cline{5-5}
Strategy          & 50      & 30   		& 20  		& 15    \\ 
\hline
Random            & 50.13	&  48.52	&  53.93	&  63.04\\
Area Split        & 49.93	&  51.02 	&  53.83	&  59.58\\
W Shape           & 51.06        &  50.05    &  50.41    &  51.20     \\
Greedy            & 47.79   &  51.19	&  53.11	&  47.90\\
Monte Carlo       & 48.02	&  53.17	&  55.90	&  57.97\\
Adaptive + MC     & 45.42	&  50.28	&  48.93	&  50.39\\
Adaptive + Greedy & 48.25   &  46.79 	&  48.96 	&  49.51
\end{tabular}
\end{table}

However, is not only important to consider the number of samples, but also the path followed by each exploration strategy. Table~\ref{tab:dist_samples} shows the average path length for different strategies. Although the adaptive sampling strategies produce $10\%$ to $30\%$ longer paths per sample, they require significantly fewer samples to produce the same quality model as their non-adaptive counterparts. Thus, the results summarised in Tables \ref{tab:dist_samples} and \ref{tab:RMSE_samples} indicate that the most efficient ways to perform Kriging-based robotic exploration use adaptive sampling strategies.

\begin{table}
\caption{Distance travelled [m] per number of samples}
\label{tab:dist_samples}
\centering
\begin{tabular}{lcccc}
                  & \multicolumn{3}{c}{\#Samples}               &       \\
Strategy          & 50                           & 30   & 20   & 15    \\ 
\hline
Random            & \hphantom{0}944	& \hphantom{0}670  & \hphantom{0}633  & \hphantom{0}450   \\
Area Split        & 1054            & \hphantom{0}820  & \hphantom{0}635  & \hphantom{0}484   \\
W Shape           & \hphantom{0}408                &  \hphantom{0}405     &  \hphantom{0}391    &   \hphantom{0}392    \\
Greedy            & 4937            & 3218 & 1967 & 1728  \\
Monte Carlo       & 3817            & 2521 & 1660 & 1238  \\
Adaptive + MC     & 1478            & \hphantom{0}841  & \hphantom{0}574  & \hphantom{0}535   \\
Adaptive + Greedy & 1317            & \hphantom{0}900  & \hphantom{0}715  & \hphantom{0}502  
\end{tabular}
\end{table}